\documentclass[lettersize,journal]{IEEEtran}
\IEEEoverridecommandlockouts
\usepackage{bbm}
\usepackage{amsfonts}
\usepackage[dvips]{graphicx}
\usepackage{times}
\usepackage{cite}
\usepackage{amsmath}
\usepackage{array}
\usepackage{amssymb}

\usepackage{stfloats}
\usepackage{graphicx}
\usepackage{footnote}
\usepackage{booktabs}
\usepackage{array}
\usepackage{algorithm}
\usepackage{subeqnarray}
\usepackage{cases}
\usepackage{threeparttable}
\usepackage{color}
\usepackage{hyperref}
\usepackage{epstopdf}
\usepackage{algpseudocode}
\usepackage{bm}
\usepackage{multirow}
\usepackage[labelformat=simple]{subcaption}
\usepackage{adjustbox}

\usepackage{enumitem}

\usepackage[normalem]{ulem}
\useunder{\uline}{\ul}{}

\makeatletter
\newcommand\fs@betterruled{%
  \def\@fs@cfont{\bfseries}\let\@fs@capt\floatc@ruled
  \def\@fs@pre{\vspace*{5pt}\hrule height.8pt depth0pt \kern2pt}%
  \def\@fs@post{\kern2pt\hrule\relax}%
  \def\@fs@mid{\kern2pt\hrule\kern2pt}%
  \let\@fs@iftopcapt\iftrue}
\floatstyle{betterruled}
\restylefloat{algorithm}
\makeatother

 \def\BibTeX{{\rm B\kern-.05em{\sc i\kern-.025em b}\kern-.08em T\kern-.1667em\lower.7ex\hbox{E}\kern-.125emX}}

\newlength{\bottomMargin}
\allowdisplaybreaks

\begin{document}
\title{Spectrum Prediction in the Fractional Fourier Domain with Adaptive Filtering}


\author{Yanghao~Qin,
        Bo~Zhou,~\IEEEmembership{Member,~IEEE,}
        Guangliang~Pan,~\IEEEmembership{Member,~IEEE,}
        Qihui~Wu,~\IEEEmembership{Fellow,~IEEE,}
        and~Meixia~Tao,~\IEEEmembership{Fellow,~IEEE}
\thanks{This research was supported in part by the National Natural Science Foundation of China (NSFC) under Grants 62231015 and 62201255. The work by M. Tao was supported by the NSFC under Grant 62125108.

 Y.~Qin, B.~Zhou and Q.~Wu are with the College of Electronic and Information Engineering, Nanjing University of Aeronautics and Astronautics, Nanjing, China.
 G.~Pan is with the College of Internet of Things, Nanjing University of Posts and Telecommunications, Nanjing, China. 
M.~Tao is with the College of Electronic Engineering, Shanghai Jiao Tong University, Shanghai, China.
Email: \{qinyanghao, b.zhou, wuqihui\}@nuaa.edu.cn, glpan@njupt.edu.cn, mxtao@sjtu.edu.cn.
}
}

\maketitle

\begin{abstract}
Accurate spectrum prediction is crucial for dynamic spectrum access (DSA) and resource allocation. 
However, due to the unique characteristics of spectrum data, existing methods based on the time or frequency domain often struggle to separate predictable patterns from noise.
To address this, we propose the Spectral Fractional Filtering and Prediction (SFFP) framework. 
SFFP first employs an adaptive fractional Fourier transform (FrFT) module to transform spectrum data into a suitable fractional Fourier domain, enhancing the separability of predictable trends from noise. 
Subsequently, an adaptive Filter module selectively suppresses noise while preserving critical predictive features within this domain. 
Finally, a prediction module, leveraging a complex-valued neural network, learns and forecasts these filtered trend components. 
Experiments on real-world spectrum data show that the SFFP outperforms leading spectrum and general forecasting methods.
\end{abstract}

 \section{Introduction}
With the proliferation of smart devices and diverse wireless services, electromagnetic spectrum resources have become increasingly constrained.
To address this challenge, dynamic spectrum access (DSA) and Cognitive Radio (CR) systems aim to improve spectrum efficiency by enabling opportunistic access to available frequency bands\cite{Tashman2021}.
Spectrum prediction plays an essential role in these systems, as it forecasts future spectrum availability based on historical data \cite{8031332}.
Accurate predictions are crucial for dynamic spectrum access and resource allocation, and they also support applications such as predictive power control, proactive frequency band switching, and spectrum anomaly detection \cite{Gao2020,10.1145/3323679.3326527}.

Conventional spectrum prediction methods typically employ statistical learning and machine learning techniques, including Autoregressive Integrated Moving Average (ARIMA), Hidden Markov Model (HMM) and Support Vector Machine (SVM).
These methods often rely on specific statistical models or require manual feature engineering, which may limit their adaptability in complex environments.
In contrast, deep learning (DL) methods have gained attention due to their ability to automatically extract features from complex and high-dimensional data \cite{10906611}.
For instance, the work in \cite{Yu2020} studies spectrum availability prediction in CR networks  using a hierarchical architecture based on Convolutional Neural Networks (CNN) and Gated Recurrent Units Networks (GRU).
Reference \cite{Gao2021} proposes a joint multi-channel and multi-step spectrum prediction algorithm that combines Long Short-Term Memory (LSTM), Seq-to-Seq modeling, and attention mechanisms.
In \cite{Pan2023}, the authors propose an Autoformer-based model with a Channel-Spatial Attention Module (Autoformer-CSA) to enhance long-term spectrum prediction.
However, most of these DL-based methods focus primarily on extracting features from the time domain, often overlooking valuable information available in the frequency domain.

Spectrum prediction can be viewed as a specific task of time series forecasting, where methods typically focus on either time or frequency domain feature extraction \cite{Wu2022, Zeng2023,Zhou2022,Zhou2022a,Xu2023}.
In the time domain, many studies decompose time series into trend, seasonality, and residual components, where trend and seasonality are considered predictable components, while the residual is treated as non-predictive noise \cite{Wu2022,Zhou2022,Zhou2022a,Zeng2023}.
Meanwhile, frequency-domain methods often apply Fourier transforms (FT) to highlight dominant periodic components (particularly low-frequency components), and may use low-pass filters to suppress high-frequency variations \cite{Zhou2022a,Xu2023}.
These methods have proven highly effective in domains such as electricity and weather forecasting, where the underlying data typically exhibit strong periodicity and clear trends \cite{qiu2024tfb}.
However, by analyzing a real‐world spectrum received signal strength (RSS) dataset (see Sec. \ref{sec:2-b}), we find that the RSS data exhibits a strong trend but weak periodicity.
Consequently, it is difficult to extract clear seasonality through decomposition or to highlight periodic structures using FT. 
Therefore, it remains unclear how to transform spectrum data and extract features in alternative domains considering its unique characteristics for accurate spectrum prediction.
\textcolor{black}{The fractional Fourier transform (FrFT) is a classical signal processing tool performing rotation in the time-frequency plane \cite{Kut1997}, and has been applied to time series forecasting \cite{Koc2022} and recently formulated as a trainable neural network component \cite{Koc2024}.}

In this paper, we propose a Spectral Fractional Filtering and Prediction framework (SFFP) for spectrum prediction. 
This framework employs an adaptive FrFT to map spectrum data into a suitable domain that facilitates both effective noise suppression and accurate trend prediction.
Specifically, it comprises three key components.
First, the FrFT module adaptively transforms the spectrum data into an optimal fractional Fourier domain where predictable trend and unpredictable noise components are more effectively separated than in the original time or frequency domains.
Then, the Filter module dynamically suppresses noise while preserving critical predictive features in the fractional Fourier domain.
Finally, the prediction module leverages a complex-valued neural network to learn and forecast the predictable trend components in the fractional Fourier domain.
Experimental results on real-world spectrum data show that our proposed SFFP outperforms leading spectrum prediction and general time-series forecasting methods.

\section{Problem Formulation and Spectrum Data Analytics}\label{sec:2}
\subsection{Problem Formulation}
We consider a spectrum monitoring system where distributed spectrum sensors monitor spectrum usage across a region of interest. 
These sensors collect RSS measurements across the $F$ discrete frequency bands and transmit them to the monitoring center.
We consider a discrete-time system indexed by $t=\{1, 2, 3,\cdots\}$. 
Our goal is to predict the RSS values for the next $\alpha$ time slots across all $F$ bands  based on the historical observations from the previous $M$ slots. 
Without loss of generality, we consider a specific point of interest and let $x_{f,t}$ denote its RSS value observed at time slot $t$ over frequency band $f$. 
We define the historical input as $\mathbf{X}_{\text{input}}=\{x_{f,t} \mid f \in \{1,\cdots, F\}, t \in \{1,\cdots, M\}\}$ and the predicted output as $\mathbf{{X}_{\text{pred}}} =\{x_{f,t} \mid f \in \{1,\cdots, F\}, t \in \{M+1,\cdots, M+P\}\}$.

We formulate this prediction task as learning a model $g: \mathbb{R}^{M \times F} \to \mathbb{R}^{P \times F}$ such that $\mathbf{{X}_{\text{pred}}}$ approximates the true future values $ \mathbf{X}_{\text{true}}$:
\begin{equation}
    \mathbf{{X}_{\text{pred}}}=g(\mathbf{{X}_{\text{input}}},\theta),
\end{equation}
with $\theta$ representing the trainable parameters.
The model can be trained by minimizing the mean squared error (MSE) loss between the predicted RSS values $\mathbf{{X}_{\text{pred}}}$ and the ground truth future values ${{X}_{\text{true}}} \in \mathbb{R}^{P \times F}$, given by:
\begin{equation} \label{eqn:loss_function}
\mathcal{L}(\theta) = \frac{1}{PF} \|\mathbf{X}_{\text{pred}} - \mathbf{X}_{\text{true}}\|_2^2.
\end{equation}

Through the training process, we aim to enable the model to learn temporal patterns in the RSS time series for accurate future signal strength predictions. 

\subsection{Spectrum Data Analytics}\label{sec:2-b}
\begin{figure}[!t]
    \centering
    \begin{subfigure}[b]{0.45\linewidth}
        \centering
        \includegraphics[width=\linewidth]{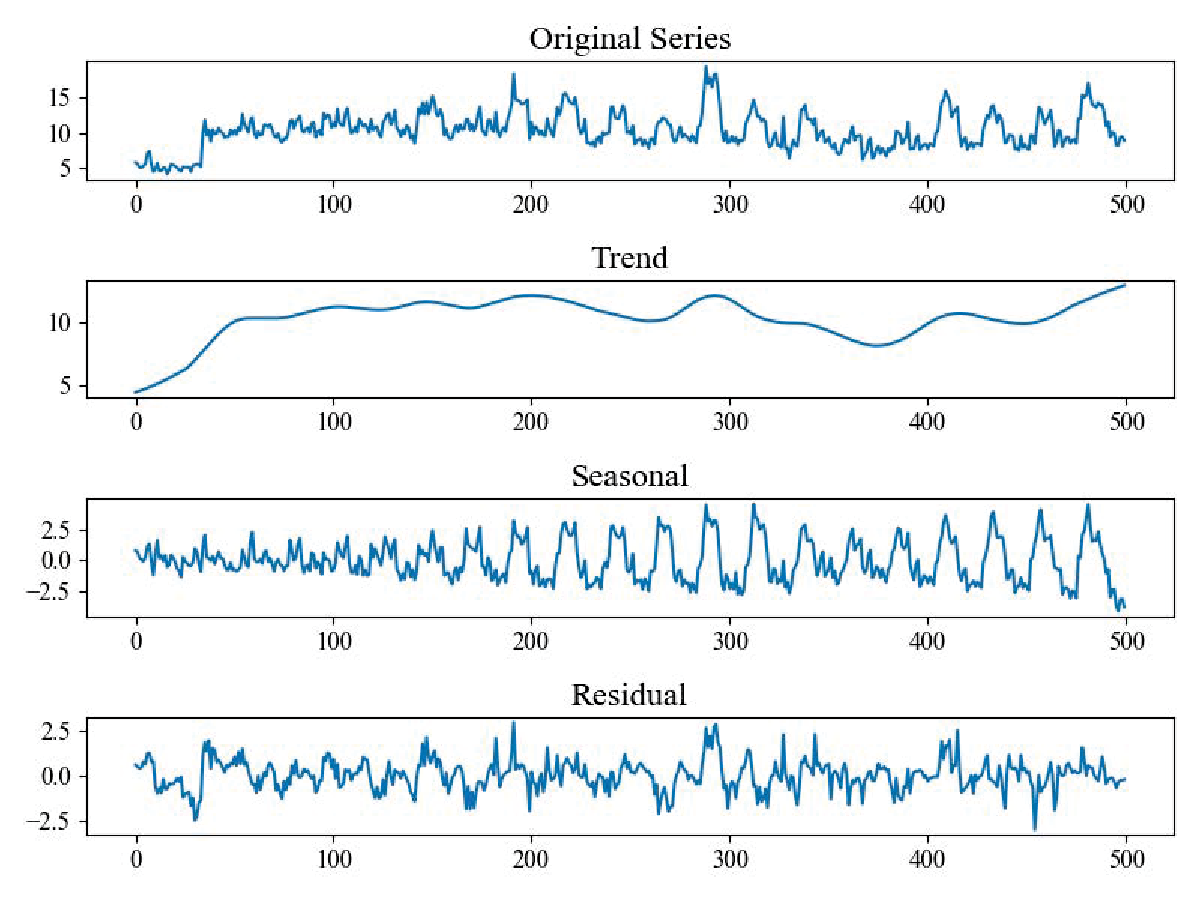}
        \caption{ETTh1 (Loess)}
        \label{fig:stl_eh1}
    \end{subfigure}
    \hfill
    \begin{subfigure}[b]{0.45\linewidth}
        \centering
        \includegraphics[width=\linewidth]{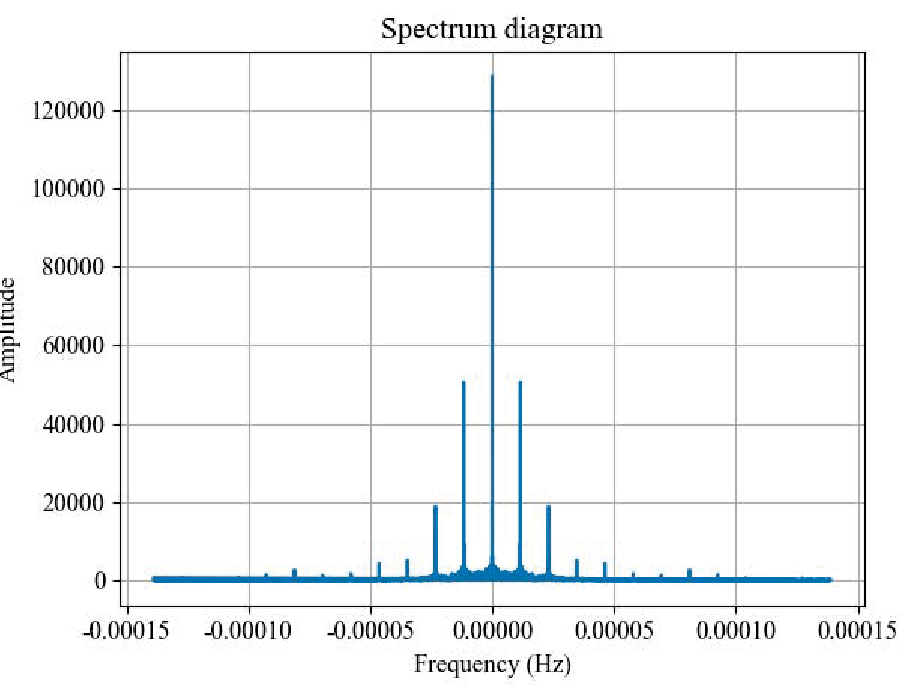}
        \caption{ETTh1 (FT)}
        \label{fig:fft_eh1}
    \end{subfigure}
    
    
    \begin{subfigure}[b]{0.45\linewidth}
        \centering
        \includegraphics[width=\linewidth]{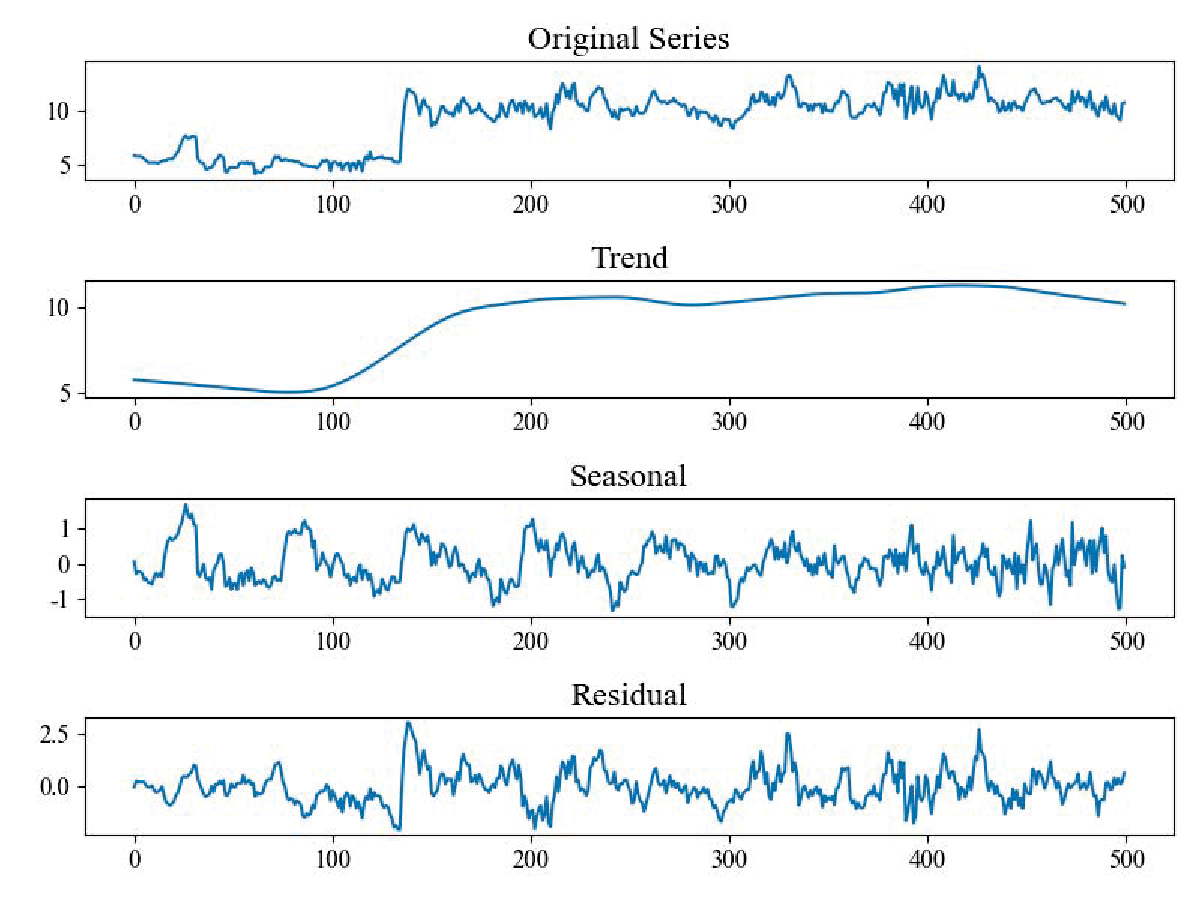}
        \caption{ETTm1 (Loess)}
        \label{fig:stl_em1}
    \end{subfigure}
    \hfill
    \begin{subfigure}[b]{0.45\linewidth}
        \centering
        \includegraphics[width=\linewidth]{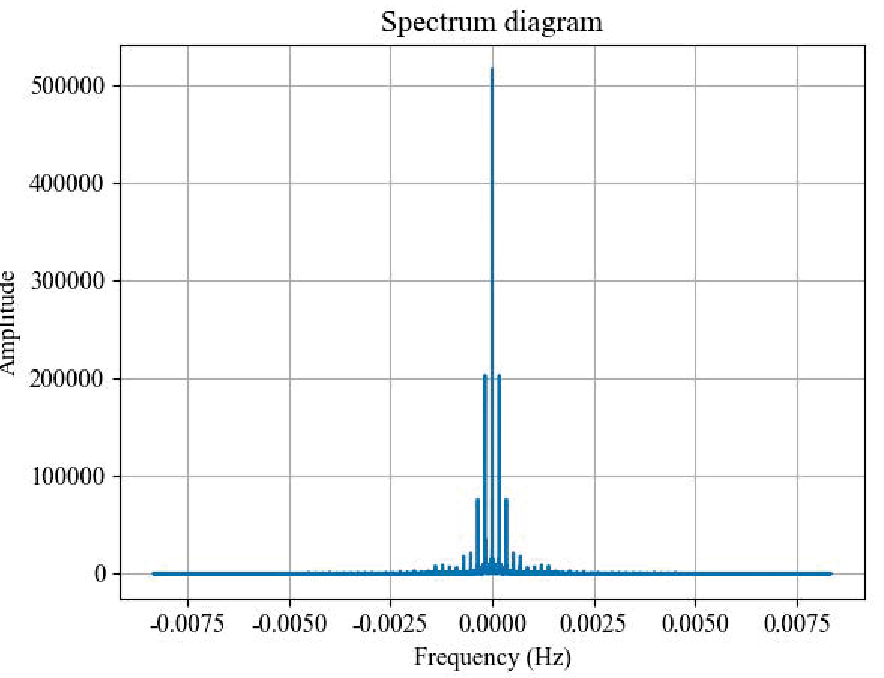}
        \caption{ETTm1 (FT)}
        \label{fig:fft_em1}
    \end{subfigure}
    
    
    \begin{subfigure}[b]{0.45\linewidth}
        \centering
        \includegraphics[width=\linewidth]{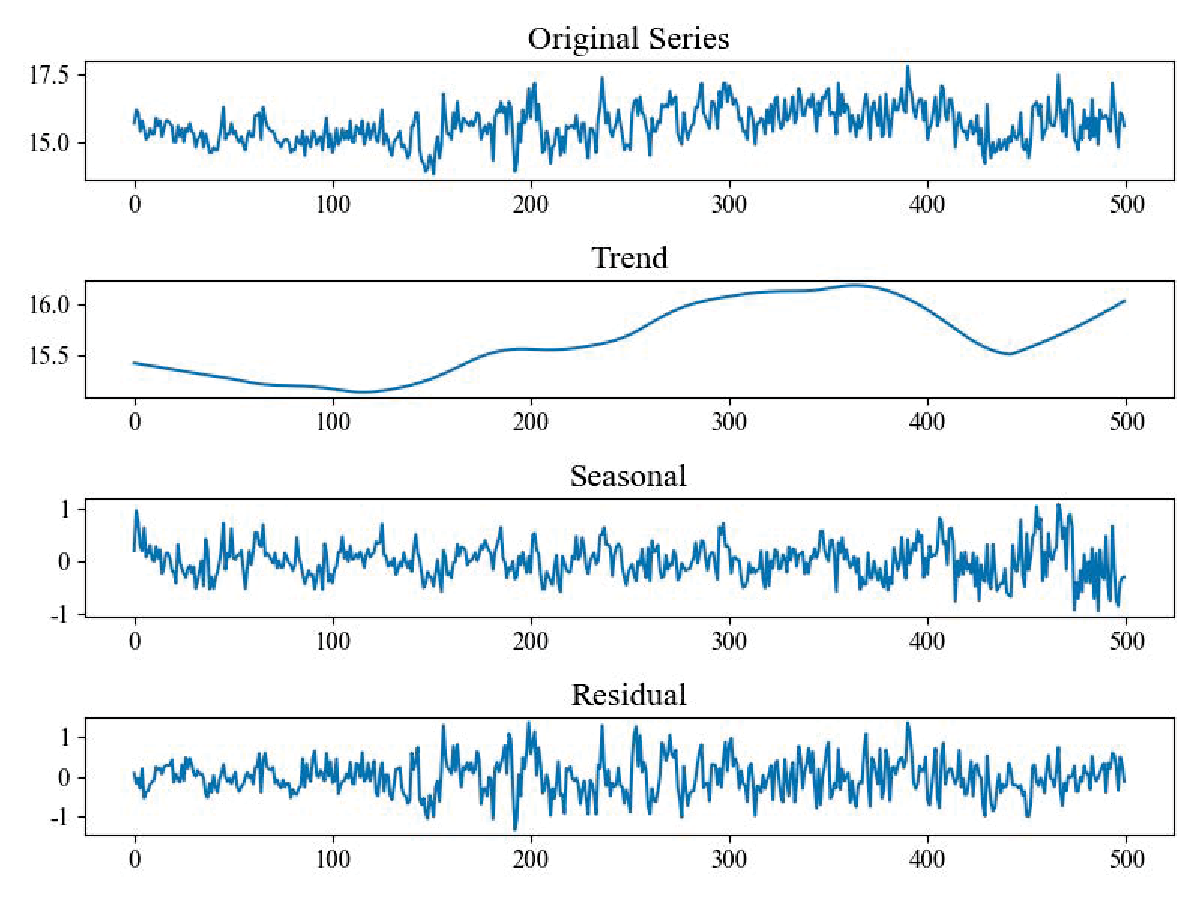}
        \caption{RSS (Loess)}
        \label{fig:stl_r3}
    \end{subfigure}
    \hfill
    \begin{subfigure}[b]{0.45\linewidth}
        \centering
        \includegraphics[width=\linewidth]{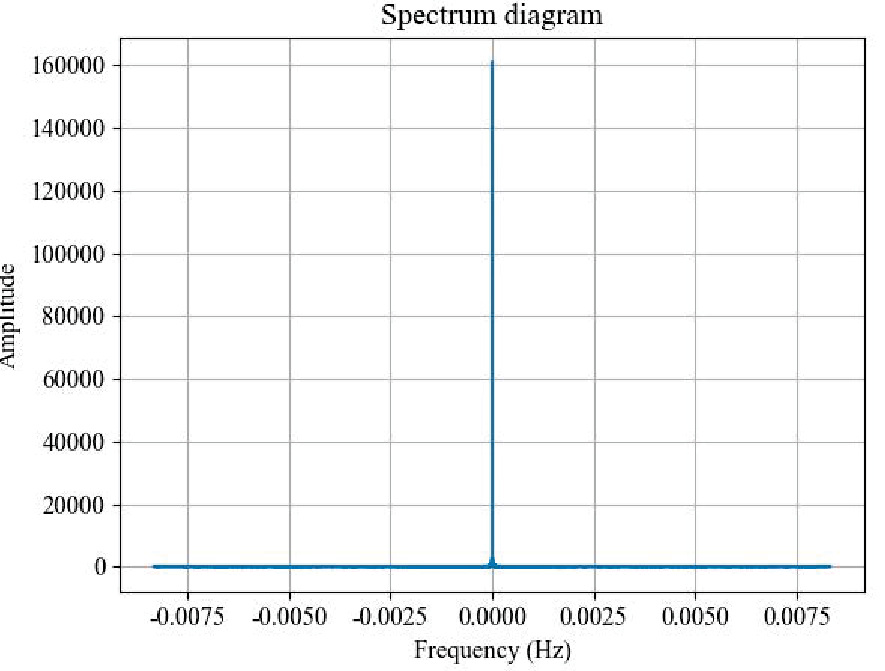}
        \caption{RSS (FT)}
        \label{fig:fft_r3}
    \end{subfigure}
    
    \caption{Time series decomposition and frequency spectrum for ETTh1, ETTm1, and RSS datasets.}
    \label{fig:time analysis results}
\vspace{-0.5cm}
\end{figure}

We analyze a real-world spectrum RSS dataset (dataset \textit{outdoor}  described in Sec. \ref{sec:dataset}) and two widely used datasets in general time series forecasting \cite{qiu2024tfb} : ETTh1 and ETTm1.
For each time series in the three datasets, we use the Loess method\cite{cleveland1990} for time-domain decomposition into trend, seasonality, and residual components and apply FT to obtain the frequency-domain representation.
From Fig. \ref{fig:stl_eh1} and Fig. \ref{fig:stl_em1}, we find that ETTh1 and ETTm1 can be effectively decomposed into clear seasonality and trend.  
Their strong periodicity is further confirmed by distinct spectral peaks in Fig. \ref{fig:fft_eh1} and Fig. \ref{fig:fft_em1}.
In contrast, for the RSS data, although Loess identifies a clear trend, the decomposed seasonal component in Fig. \ref{fig:stl_r3} lacks distinct periodic patterns in the time domain and its frequency-domain representation under FT in Fig. \ref{fig:fft_r3} shows significant spectral overlap. 

This comparison highlights a key characteristic of our real-world RSS data: it exhibits a strong trend but weak or ambiguous periodicity, unlike typical time series data.
Since existing time-series forecasting methods depend heavily on periodicity for feature extraction and learning,  they may not be directly applicable to spectrum prediction.
Therefore, for RSS data, it is more appropriate to focus on extracting the predictable trend components, rather than its weak periodicity.
Furthermore, as the significant spectral overlap under FT complicates the separation of predictive components from noise, we need to find a suitable time-frequency transformation that can alleviate this overlap and effectively suppress non-predictive noise for accurate forecasting.

\section{Proposed SFFP for Spectrum Prediction}
In this section, we propose the SFFP framework, which comprises three core components: an adaptive FrFT module, a Filter module, and a prediction module. We first introduce the use of FrFT for spectrum data, and then present the details of each module of SFFP.
\subsection{FrFT for Spectrum Data}
\textcolor{black}{The FrFT is a parameterized transformation and a generalization of the conventional FT, which introduces a real-valued order parameter $\alpha$ that allows continuous transformation between the time and frequency domains.}
 For a signal $x(t)$, its $\alpha$-order FrFT is given by \cite{Kut1997}: 
\begin{equation}\label{eqn:4}
    X_\alpha(u)=\int_{-\infty }^{+\infty }{{{K}_{\alpha}}(t,u)x(t)dt},
\end{equation}
where $ K_{\alpha}(t,u) = \sqrt{1 - i\cot \phi} \exp(i\pi ( u^2 \cot \phi - 2ut \csc \phi + t^2 \cot \phi))$ is the kernel function 
with $\phi = \frac{\alpha\pi}{2}$ and $\alpha \ne 2n$, $n \in \mathbb{Z}$. 
When $\alpha = 4n$ or $\alpha = 4n \pm 2$, the kernel simplifies to $K_\alpha(t, u) = \delta(u - t)$ or $K_\alpha(t, u) = \delta(u + t)$, respectively, and $X_\alpha(u)$ reduces to $x(t)$ or its time-inverted version $x(-t)$. 
When $\alpha = 4n+1$, FrFT simplifies to the standard FT:
$    X_{4n+1}(u) = \int_{-\infty}^{+\infty} e^{-i2\pi ut} x(t) \, dt = \left\{ \mathcal{F}(x(t)) \right\}(u),$
while $\alpha = 4n-1$ corresponds to the inverse FT.
The FrFT is periodic with respect to $\alpha$ with a period of 4, and it is sufficient to consider $\alpha \in [-2, 2]$.
It continuously interpolates between the signal ($\alpha = 0$), its FT ($\alpha= 1$), the time-inverted signal ($\alpha= \pm 2$), and the inverse FT ($\alpha= -1$). 
\textcolor{black}{Geometrically, the FrFT corresponds to a rotation in the time-frequency plane by angle $\phi$, projecting the signal onto an intermediate domain from a continuum of infinitely many domains between time and frequency \cite{Koc2022}.}
\textcolor{black}{Compared to the conventional FT, the FrFT can provide improved separability by selecting an appropriate $\alpha$, leading to enhanced noise suppression and more effective feature extraction for spectrum data.}

\begin{figure}[!t]
    \centering
    \begin{subfigure}[b]{0.45\linewidth}
        \centering
        \includegraphics[width=\linewidth]{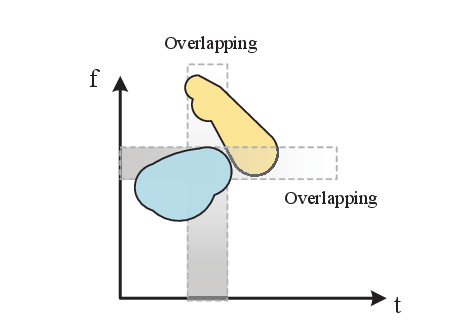}
        \caption{\centering{RSS data  using FT}}
        \label{fig:sub1}
    \end{subfigure}
    \hspace{0.05\linewidth} %
    \begin{subfigure}[b]{0.45\linewidth}
        \centering
        \includegraphics[width=\linewidth]{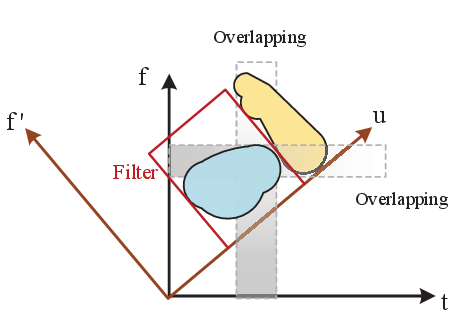}
        \caption{RSS data using  FrFT}
        \label{fig:sub2}
    \end{subfigure}
    \caption{Conceptual illustration of  FT and FrFT representations of the RSS data, where blue and yellow  regions indicate predictable and non-predictive components, respectively.}
    \label{fig:overall}
\vspace{-0.5cm} 
\end{figure}

With a proper $\alpha$, we may transform the RSS data into a fractional Fourier domain where its predictable components and non-predictive noise are more distinctly separated than in the original time or frequency domains \cite{Kut1997}.
\textcolor{black}{This allows filtering in the FrFT domain to more effectively suppress noise while preserving predictive patterns.}
Fig.~\ref{fig:overall} visualizes this concept using FT and FrFT representations of RSS data.
As discussed in Sec. \ref{sec:2-b}, the FT representation often exhibits significant spectral overlaps in the frequency domain, which hinders effective noise suppression (Fig.~\ref{fig:sub1}).
In contrast, Fig.~\ref{fig:sub2} conceptually illustrates how the FrFT with a proper $\alpha$, could yield a time-frequency representation with improved separation of predictable and non-predictive components.

\begin{figure*}[!t] 
\begin{centering} 
\includegraphics[width=0.85\textwidth]{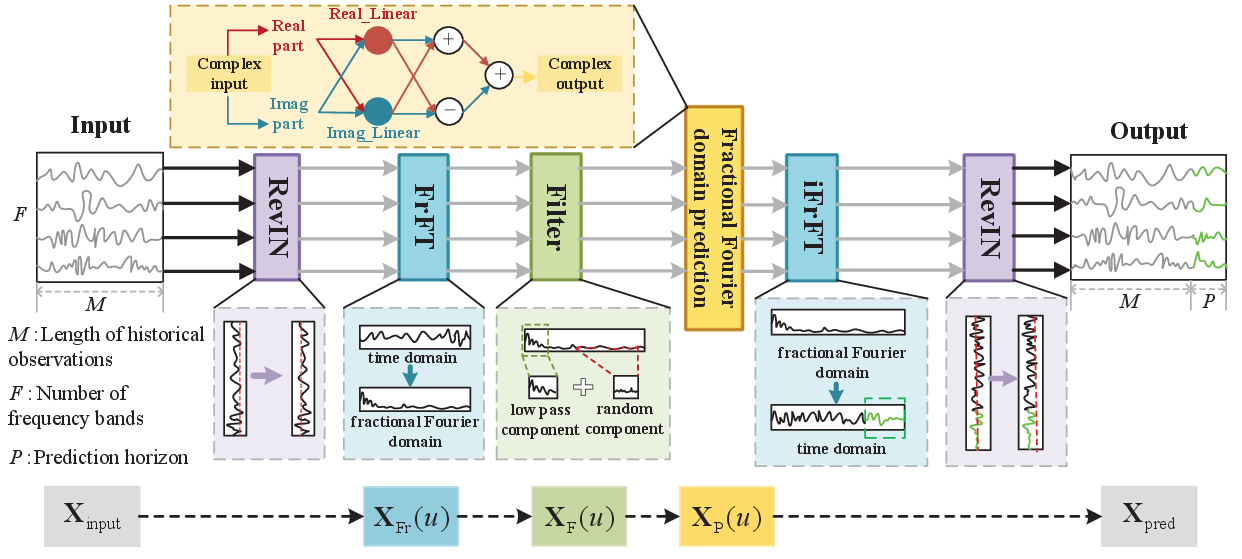} \caption{The structure of our proposed SFFP model.}
\label{fig:system} 
\end{centering} 
\vspace{-0.5cm} 
\end{figure*}

\subsection{SFFP Structure}
We present our proposed SFFP in Fig.~\ref{fig:system}.
Firstly, we propose a FrFT module which transforms the input RSS time series $\mathbf{X}_{\text{input}}$ into the fractional Fourier domain with an adaptive parameter $\alpha$. 
In the chosen fractional Fourier domain, we design a Filter module to suppress noise from the transformed data and focus on components that are useful for prediction. 
Next, we propose the fractional Fourier domain prediction module using a complex-valued linear network to learn and forecast predictable trend features within this domain.
Finally, we use an inverse FrFT (iFrFT) module which maps the predictions from the fractional Fourier domain back to the time domain, producing the final output $\mathbf{X}_{\text{pred}}$.
Additionally, we integrate an auxiliary normalization layer, Reversible Instance Normalization (RevIN) \cite{Kim2021}, into the framework to improve the model's robustness against distributional shifts.

\textbf{FrFT module:} 
\textcolor{black}{The FrFT module casts the selection of the fractional order $\alpha$ of FrFT as a learning task and learns the optimal $\alpha$ from data using the technique proposed in \cite{Koc2024}, which efficiently yields a suitable fractional Fourier-domain representation of spectrum data.}
Unlike conventional fixed-order FrFT that requires manual parameter tuning, our method directly learns the fractional order $\alpha$ by minimizing the prediction loss $\mathcal{L}$ in \eqref{eqn:loss_function} using gradient descent:
\begin{equation}\label{eq:parameter_update}
    \alpha^{(k+1)} = \alpha^{(k)} - \eta \frac{\partial \mathcal{L}}{\partial \alpha},
\end{equation}
where $\eta$ denotes the learning rate, $\alpha^{(k)}$ denotes the value of $\alpha$ at iteration $k$, and $\frac{\partial \mathcal{L}}{\partial \alpha}$ can be obtained following a similar approach as in \cite{Koc2024}.
By integrating this differentiable FrFT operator into standard DL frameworks, the module autonomously finds an appropriate RSS data representation, $\mathbf X_{\text{Fr}}(u)$, where the predictable components can be more effectively separated from noise. 

\textbf{Filter module:} 
The Filter module is proposed to suppress unpredictable noise in the learned fractional Fourier domain while focusing on components that are useful for predictions.
Inspired by the FiLM model \cite{Zhou2022}, our Filter module employs a hybrid filtering strategy operating on the complex-valued representation $\mathbf X_{\text{Fr}}(u)$.
This strategy comprises low-pass filtering and random high fractional-frequency sampling.
Specifically, we argue that $\mathbf X_{\text{Fr}}(u)$ likely concentrates most of the predictable components in low fractional-frequency regions, which can be captured using low-pass filtering.
Meanwhile, we use random high fractional-frequency sampling to retain high fractional-frequency components in $\mathbf X_{\text{Fr}}(u)$, which may contribute short-term information for prediction.
To adaptively balance the contributions of different fractional-frequency components, we introduce a learnable weight matrix $\mathbf{W}$.
Thus, the output of the Filter module is given by:
\begin{equation}
    \mathbf{X}_{\text{F}}(u) = \mathbf{W} \odot {f}_{\text{hybrid}}(\mathbf X_{\text{Fr}}(u)),
\end{equation}
where ${f}_{\text{hybrid}}(\cdot)$ represents the hybrid filtering strategy and $\odot$ denotes the element-wise operation. The weight matrix $\mathbf{W}$ is optimized during training to minimize the prediction error in \eqref{eqn:loss_function}, enabling the model to refine the selection of relevant features for predictions.
Note that the output of the Filter module is still complex valued, denoted as $\mathbf{X}_{\text{F}}(u) = \mathbf{X}_{\text{real}}(u) + j\mathbf{X}_{\text{imag}}(u)$.

\textbf{Fractional Fourier domain prediction module:} The fractional Fourier domain prediction module is designed to extract predictive features, particularly trend, and infer future RSS values by operating in the learned fractional Fourier domain.
This module uses a complex-valued linear transformation, where the learnable parameters (the slopes and biases) are complex variables, to process the complex-valued input $\mathbf{X}_{\text{F}}(u)$. 
Specifically, the real and imaginary parts of the module's output are given by:
\begin{equation}
\left\{ \begin{array}{l}
\operatorname{\mathbf{Re}\_o} = \text{L\_real}(\mathbf{X}_{\text{real}}(u)) - \text{L\_imag}(\mathbf{X}_{\text{imag}}(u)), \\ 
\operatorname{\mathbf{Im}\_o} = \text{L\_imag}(\mathbf{X}_{\text{imag}}(u)) + \text{L\_real}(\mathbf{X}_{\text{real}}(u)),
\end{array} \right.
\end{equation}
where $\text{L\_real}(\cdot)$ and $\text{L\_imag}(\cdot)$ are learnable real-valued linear layers. The use of linear models is suitable for extracting trend features, as commonly done in general time series forecasting \cite{Zeng2023,Xu2023}.
Finally, we transform the complex representation $\mathbf X_{\text{P}}(u)=\operatorname{\mathbf{Re}\_o}+j\operatorname{\mathbf{Im}\_o}$  in the fractional Fourier domain back to the time domain through iFrFT to obtain the time-domain RSS prediction  $\mathbf X_{\text{pred}}$.
\textcolor{black}{In summary, the SFFP model processes the input $\mathbf X_{\text{input}}$ by applying the FrFT module to obtain $\mathbf X_{\text{Fr}}(u)$, the Filter module to produce refined features $\mathbf X_{\text{F}}(u)$, the prediction module to generate fractional Fourier-domain prediction $\mathbf X_{\text{P}}(u)$, and finally the iFrFT to obtain the time-domain forecast $\mathbf X_{\text{pred}}$.}

\section{Experimental Results and Discussions}
\subsection{Experiment Setup}\label{sec:dataset}
\textbf{Dataset:} 
\textcolor{black}{
We use three datasets of RSS measurements. The first two, referred to as the \textit{outdoor} and \textit{indoor} datasets, are obtained via the ElectroSense open API\footnote{https://electrosense.org/} and contain measurements collected in Madrid, Spain, from 1 June to 8 June, 2021, with a 1-minute sampling interval. The frequency range is 600–612 MHz with a 2 MHz bandwidth. The third dataset, referred to as the \textit{campus} dataset, was collected at Nanjing University of Aeronautics and Astronautics, Nanjing, China, from 9am to 12pm on September 23, 2022, with a 1-second sampling interval. Its frequency range is 83.4–115.4 MHz with a 1 MHz bandwidth.}

\textbf{Baselines:}
We compare our proposed SFFP with Autoformer-CSA \cite{Pan2023}, a state-of-the-art model for spectrum prediction, and other leading time-series forecasting models—TimesNet \cite{Wu2022}, DLinear \cite{Zeng2023}, FEDFormer \cite{Zhou2022}, FiLM \cite{Zhou2022a}, and FITS \cite{Xu2023}.

\textbf{Additional details:} 
We use MSE and Mean Absolute Error (MAE) as evaluation metrics and split the datasets into training, validation, and test sets in an 8: 1: 1 ratio.
All experiments use an input length $M=96$ and a prediction horizon of $P=24$ unless otherwise specified.
All results are averaged over 50 independent experiments.

\begin{table}[ht]
\small
\centering
\caption{Performance comparison under different prediction horizons $P \in \{24, 48, 96\}$.}
\label{tab:all_results}
\resizebox{\linewidth}{!}{ 
\begin{tabular}{ccccccccc}
\toprule
\multirow{2}{*}{Model} & \multirow{2}{*}{$P$} & \multicolumn{2}{c}{\textit{outdoor}} & \multicolumn{2}{c}{\textit{indoor}} & \multicolumn{2}{c}{\textit{campus}} \\
\cmidrule(lr){3-4} \cmidrule(lr){5-6} \cmidrule(lr){7-8}
& & MSE & MAE & MSE & MAE & MSE & MAE \\
\midrule
\multirow{3}{*}{TimesNet} 
  & 24 & 0.6896 & 0.6589 & 0.1523 & 0.2390 & 0.7633 & 0.6804 \\
  & 48 & 0.8369 & 0.7218 & 0.1731 & 0.2585 & 0.8075 & 0.7044 \\
  & 96 & 1.1776 & 0.8624 & 0.2001 & 0.2804 & 0.8224 & 0.7167 \\
\midrule
\multirow{3}{*}{DLinear} 
  & 24 & 0.6175 & 0.6266 & 0.1512 & 0.2377 & 0.7684 & 0.6857 \\
  & 48 & \underline{0.7017} & 0.6667 & \underline{0.1627} & \underline{0.2483} & 0.7997 & 0.7030 \\
  & 96 & \underline{0.8658} & \underline{0.7410} & \underline{0.1774} & \underline{0.2604} & 0.8220 & 0.7117 \\
\midrule
\multirow{3}{*}{FEDformer} 
  & 24 & 0.6890 & 0.6612 & 0.1748 & 0.2644 & 0.7629 & 0.6788 \\
  & 48 & 0.8680 & 0.7384 & 0.1853 & 0.2725 & 0.8067 & 0.7020 \\
  & 96 & 0.9174 & 0.7683 & 0.2017 & 0.2868 & 0.8287 & 0.7116 \\
\midrule
\multirow{3}{*}{Autoformer-CSA} 
  & 24 & 0.7688 & 0.6980 & 0.1760 & 0.2689 & 0.7722 & 0.6846 \\
  & 48 & 0.8101 & 0.7148 & 0.1890 & 0.2774 & 0.7964 & 0.6995 \\
  & 96 & 0.9620 & 0.7825 & 0.2010 & 0.2841 & 0.8196 & 0.7102 \\
\midrule
\multirow{3}{*}{Film} 
  & 24 & 0.7194 & 0.6763 & 0.1752 & 0.2661 & 0.7671 & 0.6855 \\
  & 48 & 0.8026 & 0.7111 & 0.1872 & 0.2757 & 0.7970 & 0.7001 \\
  & 96 & 0.9849 & 0.7898 & 0.2011 & 0.2850 & 0.8246 & 0.7121 \\
\midrule
\multirow{3}{*}{FITS} 
  & 24 & \underline{0.6162} & \underline{0.6251} & \underline{0.1417} & \underline{0.2294} & \underline{0.7500} & \underline{0.6710} \\
  & 48 & 0.7059 & \underline{0.6660} & 0.1642 & 0.2505 & \underline{0.7836} & \underline{0.6897} \\
  & 96 & 0.8712 & 0.7423 & 0.1876 & 0.2680 & \underline{0.8096} & \underline{0.7035} \\
\midrule
\multirow{3}{*}{SFFP} 
  & 24 & \textbf{0.5798} & \textbf{0.6052} & \textbf{0.1386} & \textbf{0.2256} & \textbf{0.7319} & \textbf{0.6616} \\
  & 48 & \textbf{0.6674} & \textbf{0.6480} & \textbf{0.1536} & \textbf{0.2399} & \textbf{0.7707} & \textbf{0.6825} \\
  & 96 & \textbf{0.8256} & \textbf{0.7223} & \textbf{0.1771} & \textbf{0.2595} & \textbf{0.8031} & \textbf{0.6988} \\
\bottomrule
\end{tabular}
}
\label{table:all results}
\vspace{-0.6cm}
\end{table}

\subsection{Experiment Results}
Table \ref{table:all results} compares the prediction performance between SFFP and the baselines under different prediction horizons on \textcolor{black}{the three datasets}.
We can see that our proposed SFFP consistently outperforms all baselines.
For example, when $P=24$, on dataset \textit{outdoor}, SFFP reduces MSE by 24.58\% compared to spectrum-specific baseline Autoformer-CSA, and 5.91\% compared to FITS, the strongest general forecasting baseline in this case; 
on dataset \textit{indoor}, the reductions are 21.25\% and 2.19\%; \textcolor{black}{and on dataset \textit{campus}, they are 5.22\% and 2.41\%.}

\begin{figure}[htbp]
    \begin{minipage}[t]{0.485\linewidth}
        \centering
        \includegraphics[width=1\columnwidth]{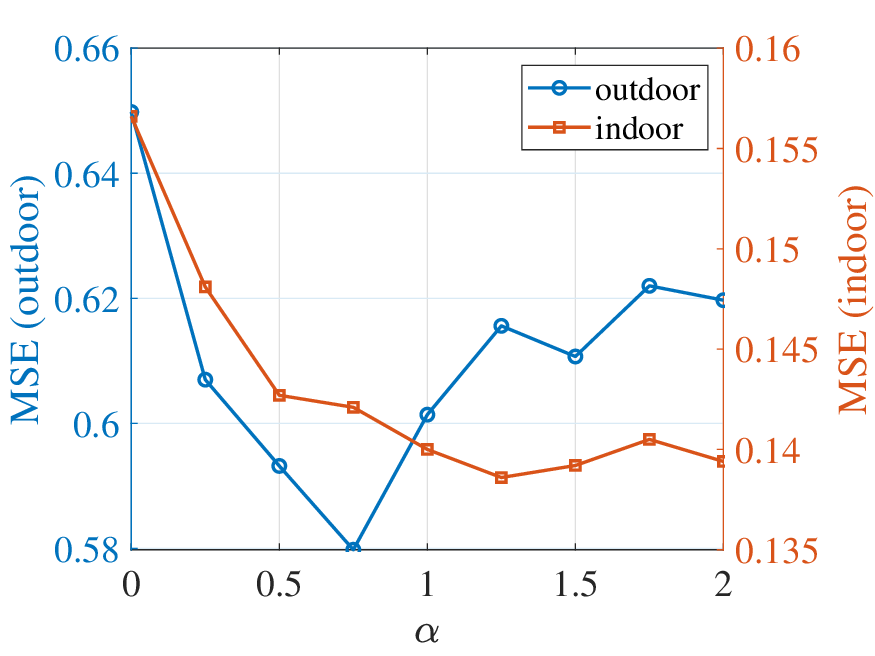}
        \caption{Effect of fractional order  $\alpha$ on MSE.\label{fig:ex2_p}}
    \end{minipage}%
    \hspace{0.1cm} 
    \begin{minipage}[t]{0.485\linewidth}
        \centering
        \includegraphics[width=1\columnwidth]{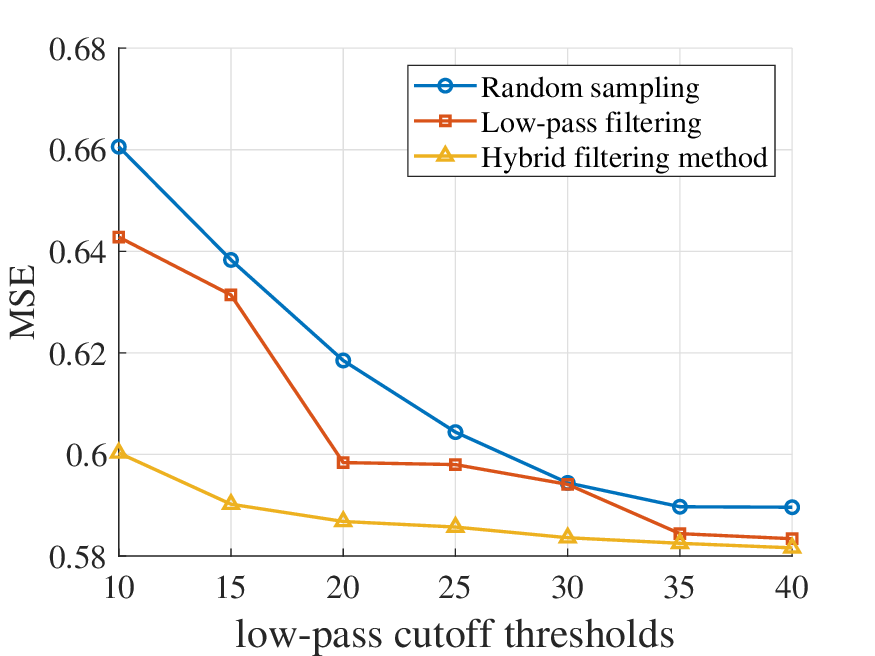}
        \caption{Comparison of filter strategies on MSE.\label{fig:ex3}}
    \end{minipage}
\vspace{-0.2cm} 
\end{figure}

Fig.~\ref{fig:ex2_p} illustrates the effect of the fractional order parameter $\alpha$ on the MSE performance of our proposed SFFP method. 
We observe that MSE is highly sensitive to $\alpha$, with the optimal values learned by our adaptive FrFT module being $\alpha=0.75$ for dataset \textit{outdoor} and $\alpha=1.25$ for dataset \textit{indoor}.
Using these learned values, the corresponding fractional Fourier-domain approaches achieve significantly lower MSE than processing directly in the standard time-domain ($\alpha=0$) and frequency-domain ($\alpha=1$), which validates the effectiveness of our adaptive fractional Fourier-domain transformation.

To further evaluate our FrFT module, Table~\ref{table:frft} presents an ablation study comparing different combinations of input transformations and prediction architectures. 
We evaluate three types of transformations: No transformation (raw time domain input), FT, and adaptive FrFT, combined with four prediction models: Transformer, CNN, LSTM, and the Linear model used in SFFP.
Consistent with Fig.~\ref{fig:ex2_p}, the adaptive FrFT outperforms both the time-domain and the FT approaches across all prediction architectures. 
Moreover, the Linear model outperforms the more complex Transformer, CNN, and LSTM models, which reveals its effectiveness in capturing dominant trend predictive patterns in RSS forecasting.

\begin{table}[ht]
\small
\centering
\caption{Performance comparison of different input transformations and predictors on dataset \textit{outdoor}.}
\label{tab:frft}
\resizebox{0.7\linewidth}{!}{ 
\begin{tabular}{cccc}
\toprule
\multirow{2}{*}{Predictor} & \multirow{2}{*}{Transformation} & \multicolumn{2}{c}{\textit{outdoor}} \\
\cmidrule(lr){3-4}
& & MSE & MAE \\
\midrule
\multirow{3}{*}{Transformer} 
  & NO & 0.8732 & 0.7392 \\
  & FFT & 0.7637 & 0.7148 \\
  & FrFT & \textbf{0.7546} & \textbf{0.6879} \\
\midrule
\multirow{3}{*}{CNN} 
  & NO & 1.6162 & 0.9962 \\
  & FFT & 0.7527 & 0.6921 \\
  & FrFT & \textbf{0.6951} & \textbf{0.6415} \\
\midrule
\multirow{3}{*}{LSTM} 
  & NO & 0.7213 & 0.6874 \\
  & FFT & 0.6814 & 0.6642 \\
  & FrFT & \textbf{0.6585} & \textbf{0.6463} \\
\midrule
\multirow{3}{*}{Linear} 
  & NO & 0.6145 & 0.6236 \\
  & FFT & 0.6155 & 0.6237 \\
  & FrFT & \textbf{0.5834} & \textbf{0.6067} \\
\bottomrule
\end{tabular}
}
\label{table:frft}
\vspace{-0.6cm} 
\end{table}




Fig.~\ref{fig:ex3} illustrates the MSE performance with varying low-pass cutoff thresholds under three filtering strategies on dataset \textit{outdoor}, i.e., random sampling, low-pass filtering, and our proposed hybrid filtering method.
We can see that our hybrid method consistently achieves the lowest MSE, especially with smaller low-pass cutoff thresholds.
This validates the benefit of combining low-pass filtering (to capture dominant predictable components based on the threshold) with random high fractional-frequency sampling (to potentially retain useful finer details).

\section{Conclusion}
In this letter, we propose the SFFP framework for spectrum prediction. 
SFFP integrates an adaptive FrFT module to map data into an optimal fractional Fourier domain for trend-noise separation, an adaptive filter to selectively remove noise while retaining key predictive patterns, and a prediction module for predicting these processed fractional Fourier domain components.
Experiments on real-world datasets confirm SFFP's significant performance, further validating the contributions of each module to effective spectrum prediction.

\bibliographystyle{IEEEtran}
\def\baselinestretch{0.97}
\bibliography{IEEEabrv,reference}

\end{document}